# Bayesian Rose Trees

**Charles Blundell, Yee Whye Teh**
Gatsby Computational Neuroscience Unit,
University College London,
London, UK

**Katherine A. Heller**
Department of Engineering,
University of Cambridge,
Cambridge, UK

## Abstract

Hierarchical structure is ubiquitous in data across many domains. There are many hierarchical clustering methods, frequently used by domain experts, which strive to discover this structure. However, most of these methods limit discoverable hierarchies to those with binary branching structure. This limitation, while computationally convenient, is often undesirable. In this paper we explore a Bayesian hierarchical clustering algorithm that can produce trees with arbitrary branching structure at each node, known as *rose trees*. We interpret these trees as mixtures over partitions of a data set, and use a computationally efficient, greedy agglomerative algorithm to find the rose trees which have high marginal likelihood given the data. Lastly, we perform experiments which demonstrate that rose trees are better models of data than the typical binary trees returned by other hierarchical clustering algorithms.

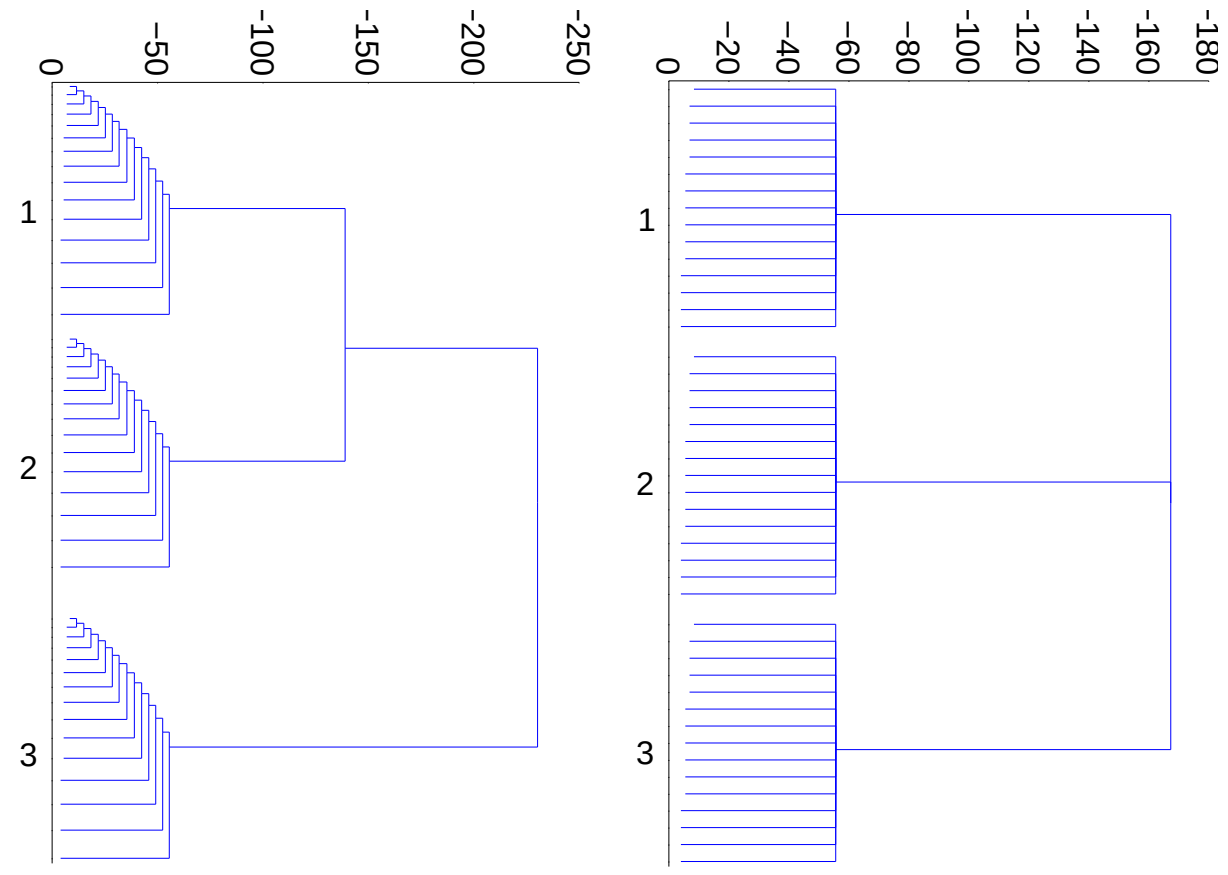

Figure 1: Bayesian hierarchical clustering (left) and Bayesian rose trees (right) on the same synthetic data set. The three groups of 15 similar data points all cluster into groups under both models. However Bayesian hierarchical clustering defines a mixture model over $3,391$ partitions whereas Bayesian rose trees defines a mixture model over just 11 partitions of the data with a higher marginal likelihood (marginal likelihood is on the horizontal axis).

## 1 Introduction

An important part of exploratory data analysis and unsupervised learning is determining what hierarchical structure, if any, exists in data. Rich hierarchies are common in data across many domains. For example, topic hierarchies are found in text processing, phylogenies in evolutionary biology, and hierarchical community structures in social networks.

Most algorithms for hierarchical clustering construct binary tree representations of data, where leaf nodes correspond to data points and internal nodes correspond to clusters. For example, traditional agglomerative linkage methods (Duda and Hart, 1973) start with each data point in its own cluster and iteratively merge the closest pair of clusters, as determined by some distance metric, together until all data belong to a single cluster. Current probabilistic and Bayesian approaches to hierarchical clustering, to be discussed later, produce binary trees as well.

This restriction of the hypothesis space to binary trees alone is undesirable in many situations. Firstly, we simply do not believe that many hierarchies in real world applications are binary. Secondly, limiting algorithms to binary trees often forces spurious structure to be hallucinated even if this structure is not supported by data, making the practitioners task of interpreting the trees more difficult. This spurious structure is also undesirable from an Occam's razor point of view. The algorithms are not returning the simplest structure supported by the data, because the simpler models which explain the data have been excluded from the hypothesis space. Figure 1(left) shows an example of such hallucinated structure returned by Bayesian hierarchical clustering (BHC) (Heller and Ghahramani, 2005). In this case cascades are used to represent the three large clusters. This is a telltale sign among probabilistic binary tree construction algorithms that the tree cannot represent the large clusters in the data properly. Ideally the tree structure should be simplified by collapsing each cascade into a single node with many children expressing the indistinguishability among the children, as in Figure 1(right).

In this paper we broaden the hypothesis space of our hierarchical clustering algorithm to include trees with arbitrary branching structure at each internal node. We refer to these as *rose trees*, as they are known in the functional programming literature (Meertens, 1988). Since there are many more options for branching structure, the space of rose trees is larger than binary trees[1], and the search for good trees is correspondingly harder.

For the sake of computational efficiency, we take a greedy agglomerative approach to constructing trees, and consider three ways in which subtrees can be merged: a join operation that creates a new node, an absorb that does not, and a collapse that removes a node (see figure 3).

We will show on multiple data sets that this algorithm produces very good trees. Not only does our method obtain more probable explanations of data, but it also results in simpler, easier to interpret hierarchies.

In Section 2 we describe our model in detail and discuss relationships with BHC. In Section 3 we describe our agglomerative construction algorithm. In Section 4 we report experimental results using Bayesian rose trees, and finally, we conclude with a discussion of previous related work in Section 5.

## 2 Rose trees, partitions and mixtures

The starting point of our approach is Bayesian hierarchical clustering (BHC) (Heller and Ghahramani, 2005), a probabilistic approach to hierarchical clustering. In BHC a tree is associated with a set of tree-consistent partitions, and interpreted as a mixture model, where each component of the mixture is represented by a subset of tree nodes, and corresponds to a partition of the data.

We start by giving proper definitions of rose trees, partitions, and our interpretation of rose trees as mixtures over partitions. A rose tree is defined recursively: $T$ is a rose tree if either $T = \{x\}$ for some data point $x$, or $T = \{T_1, \ldots, T_{n_T}\}$ where $T_i$'s are rose trees over disjoint sets of data points. In the latter case each $T_i$ is a child of $T$ and $T$ has $n_T$ children. Let leaves$(T)$ be the set of data points at the leaves of $T$.

Our concepts of partitions and mixtures over partitions are direct generalisations of the binary tree case (Heller and Ghahramani, 2005). We denote partitions using "|", for example $ab|c$ denotes a partition of the set $\{a, b, c\}$ into disjoint subsets $\{a, b\}$ and $\{c\}$. A rose tree $T$ is used to represent a structured subset $\mathbb{P}(T)$ of all partitions of some data points $\mathcal{D}$. Specifically it represents the set of partitions consistent with $T$, which can be defined recursively as follows:

$$\mathbb{P}(T) = \{\text{leaves}(T)\} \cup \left\{ \phi_1|\ldots|\phi_{n_T} : \begin{array}{c} T_i \in \text{ch}(T), \\ \phi_i \in \mathbb{P}(T_i) \end{array} \right\} \quad (1)$$

where ch$(T)$ are the children of $T$, and $\{\text{leaves}(T)\}$ represents the partition where all data points at the leaves of T are clustered together. Two examples of the sets of partitions associated with rose trees are shown in Figure 2. Roughly, each partition starts at the root of the tree, and either keeps the leaves in one cluster or partitions the leaves into the subtrees, the process repeating on each subtree. The end result is that each $\phi \in \mathbb{P}(T)$ consists of non-overlapping clusters, each of which consists of all the leaves of some subtree in $T$. We will denote these subtrees by front$_T(\phi)$ and the set of ancestors of subtrees in front$_T(\phi)$ by an$_T(\phi)$. Note that all rose trees include the complete partition $\{\text{leaves}(T)\}$ and (by recursion) the completely discriminating partition where each data point in $\mathcal{D}$ is in its own component of the partition.

We interpret a rose tree $T$ as a mixture over partitions in $\mathbb{P}(T)$ of the data points at its leaves $\mathcal{D} = \text{leaves}(T)$:

$$p(\mathcal{D}|T) = \sum_{\phi_T \in \mathbb{P}(T)} p(\phi_T) p(\mathcal{D}|\phi_T) \quad (2)$$

where $p(\phi_T)$ is the mixing proportion of partition $\phi_T$, and $p(\mathcal{D}|\phi_T)$ is the probability of data $\mathcal{D}$ given a partitioning by $\phi_T$. In general the number of partitions consistent with $T$ can be exponentially large. To make computations tractable, we define the mixture model in such a way that $p(\mathcal{D}|T)$ can be computed using dynamic programming over $T$:

$$p(\mathcal{D}|T) = \pi_T f(\mathcal{D}) + (1-\pi_T) \prod_{T_i \in \text{ch}(T)} p(\text{leaves}(T_i)|T_i) \quad (3)$$

[1]Asymptotically there are a factor of $2^{O(n)}$ more rose trees than binary trees.

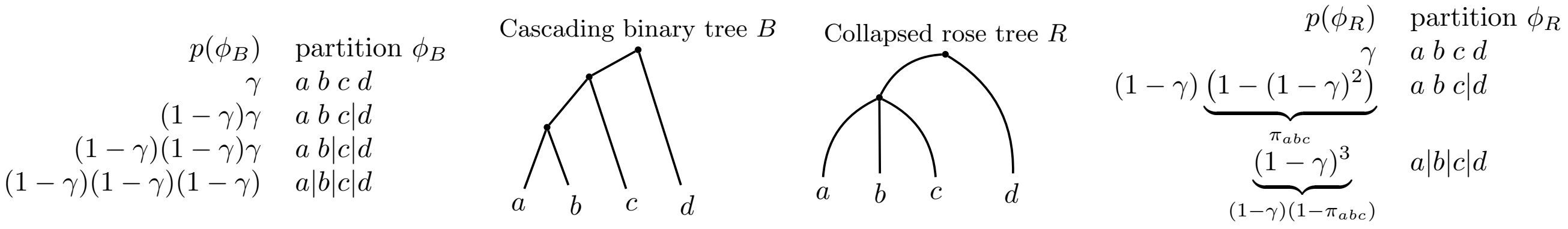

Figure 2: The BRT mixing proportions are defined such that mass associated with the partition $\{a\,b|c|d\}$ missing from the collapsed rose tree $R$ is re-assigned to the complete partition $\{a\,b\,c|d\}$.

where $f(\mathcal{D})$ is the marginal probability of the data $\mathcal{D}$ under an exponential family, with the parameters marginalized out under a conjugate prior with hyperparameters $\beta$, and $\pi_T$ is a mixing proportion. Comparing (3) and (2), we find that

$$p(\mathcal{D}|\phi_T) = \prod_{D' \in \phi_T} f(D') \tag{4}$$

$$p(\phi_T) = \prod_{A \in \mathrm{an}_T(\phi_T)} (1 - \pi_A) \prod_{S \in \mathrm{front}_T(\phi_T)} \pi_S \tag{5}$$

The probability of $\mathcal{D}$ under partition $\phi_T$ is simply the probability of each cluster $D'$ of data points in $\phi_T$ under the exponential family, while $\pi_T$ is the prior probability that the leaves under $T$ are kept in one cluster rather than subdivided by the recursive partitioning process. We define $\pi_T$ as follows:

$$\pi_T = 1 - (1-\gamma)^{n_T - 1} \tag{6}$$

where $0 \leq \gamma \leq 1$ is a hyperparameter of the model controlling the relative proportion of coarser partitions of the data as opposed to finer ones. When restricting to binary trees only, $\pi_T = \gamma$ and the model reduces to the constant $\pi$ version in Heller and Ghahramani (2005). This choice of $\pi_T$ is intimately related to our maxim that the maximum likelihood tree should be simple if the data is unstructured, and will be explained in the next two subsections.

In summary, the marginal probability of $\mathcal{D}$ under a rose tree $T$, $p(\mathcal{D}|T)$, is a mixture over the partitions consistent with $T$, with the probability of $\mathcal{D}$ under a partition $\phi \in \mathbb{P}(T)$ being a product $\prod_{D \in \phi} f(D)$ of the probabilities of clusters in $\phi$. We call our mixture a *Bayesian rose tree* (BRT) mixture model.

### 2.1 Avoiding needless cascades

In this section we explain our choice of $\pi_T$ given in equation 6. We will start with a simple situation consisting of four data points $a$, $b$, $c$ and $d$ depicted in Figure 2.

Consider the two rose trees in Figure 2 over the data points $\mathcal{D}$, consisting of $a, b, c$ and $d$. Suppose that the data points $a, b, c$ are similar to each other but are otherwise indistinguishable, i.e. they should be in just one cluster, yet are distinguishable from $d$. We should prefer the collapsed rose tree $R$ over the cascading binary tree $B$. The figure also shows the set of partitions and their mixing proportions under BHC-$\gamma$, for $B$, and under BRT, for $R$. Because the data points $a, b, c$ belong together in one cluster, we can expect the following inequalities among the marginal likelihoods of the data under the partitions (recall $p(\mathcal{D}|\phi) = \prod_{D \in \phi} f(D)$ is the likelihood of partition $\phi$):

$$p(\mathcal{D}|\{a\,b\,c|d\}) > p(\mathcal{D}|\{a\,b|c|d\}) \tag{7}$$

$$p(\mathcal{D}|\{a\,b\,c|d\}) > p(\mathcal{D}|\{a\,c|b|d\}) \tag{8}$$

$$p(\mathcal{D}|\{a\,b\,c|d\}) > p(\mathcal{D}|\{b\,c|a|d\}) \tag{9}$$

We want, where possible, the model to prefer $R$ over $B$, and so also require

$$p(\mathcal{D}|R) > p(\mathcal{D}|B) \tag{10}$$

Expanding the marginal likelihoods under $R$ and $B$ as a mixture of the likelihoods under each partition, and using the inequality among the partition likelihoods we can guarantee (10) if we set the mixing proportion $\pi_{abc}$ of the subtree $R$ with leaves $a, b, c$, to be $1 - \pi_{abc} = (1-\gamma)^2$. Here the mass of missing partitions from $B$ are re-assigned to the collapsed partition in $R$. This is shown in figure 2.

In the general case, if we have a cluster of indistinguishable data points, we can guarantee preferring a rose tree $R$ consisting of a single internal node over any binary tree if the mixing proportion of the complete partition in $R$ is the sum over the mixing proportions of all partitions consistent with the binary tree except the most discriminating partition. Fortunately, this sum turns out to be the same regardless of the structure of the binary tree (if $\pi_B = \gamma$ for each binary tree $B$), and equals:

$$\pi_R = 1 - (1-\gamma)^{n_R - 1} \tag{11}$$

where $n_R$ is the number of children of $R$.

As in BHC-DP, $\pi_T$ parameterised as in equation 11 also tends to one as the number of children becomes large but it does so much more slowly (compared to the BHC mixture proportion assignment). When $n_R = 2$, this assignment agrees with BHC-$\gamma$.

### 2.2 Relation to BHC and DP mixture models

Bayesian rose trees are a strict generalisation of BHC—if every node is restricted to have just two children we will recover BHC. Heller and Ghahramani (2005) described two parametrisation of $\pi_T$ which we shall refer to as BHC-$\gamma$ and BHC-DP. BHC-$\gamma$ sets $\pi_T = \gamma$, $\gamma$ being a fixed hyperparameter, and the BRT model we just described is a generalisation of this model. On the other hand BHC-DP sets up $\pi_T$ such that it produces a lower bound on the marginal likelihood of a corresponding Dirichlet process (DP) mixture. A similar set-up can allow BRT to produce a lower bound as well, though we will now argue that this is in fact undesirable.

Recall that the marginal probability of data under a DP mixture model is a convex combination of exponentially many terms, each of which is the probability of the data under a different partition of the data items into clusters. BHC-DP produces a lower bound on this marginal probability by including only the terms corresponding to partitions which are consistent with the constructed binary tree. A similar setting of $\pi_T$'s in BRT allows it to also produce a lower bound on the DP mixture marginal likelihood. However, BRTs generally correspond to much smaller sets of partitions than binary trees—if we replace each non-binary internal node of the rose tree with a cascade of binary nodes we will get a superset of partitions (see also Figure 1 and Section 4). This implies that the BRT lower bound will be no higher than the BHC lower bound.

The above argument obviates the use of Bayesian rose trees as an approximate inference method for DP mixtures since they correspond to smaller sets of partitions of the data. In fact our reason for using rose trees is precisely because the sets of partitions are smaller—if there is no structure in the data to support a more complex model, by Occam's Razor we should prefer a simpler model (reflected in terms of a smaller number of partitions). This view of hierarchical clustering is very different from the one expounded by Heller and Ghahramani (2005).

## 3 Greedy construction of Bayesian rose tree mixtures

We take a model selection approach to finding a rose tree structure given data. Ideally, we wish to find a rose tree $T^*$ maximising the marginal probability of the data $\mathcal{D}$:

$$T^* = \operatorname*{argmax}_T p(\mathcal{D}|T) \tag{12}$$

This is intractable since there is a super-exponential number of rose trees.

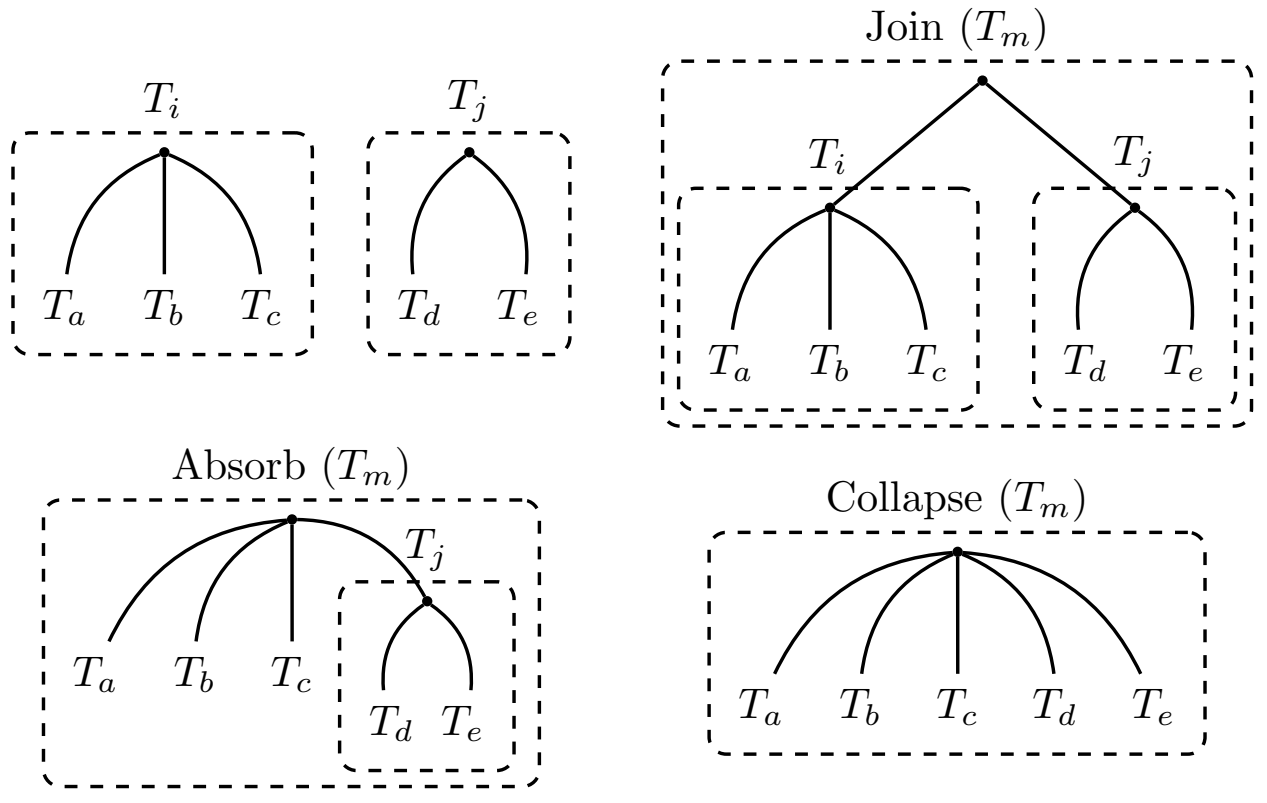

Figure 3: Merges considered during greedy search.

Instead, rose trees can be constructed in a greedy agglomerative fashion as follows. Initially every data point is assigned to its own rose tree: $T_i = \{x_i\}$ for all data points $x_i$. At each step of our algorithm we pick two rose trees $T_i$ and $T_j$ and merge them into one tree $T_m$. This procedure repeats until just one tree remains (for $n$ data points this will occur after $n-1$ merges).

To allow for nodes with more than two children, we consider three types of merges which we call a join, an absorb, and a collapse (Figure 3). In all operations the merged rose tree $T_m$ has $\text{leaves}(T_m) = \text{leaves}(T_i) \cup \text{leaves}(T_j)$, the difference being the resulting structure at the root of the merged tree. For a join, a new node is created with children $T_m = \{T_i, T_j\}$. For an absorb $T_m = \text{ch}(T_i) \cup \{T_j\}$, that is, tree $T_j$ is absorbed as a child of $T_i$. This operation is not symmetric so we also consider the converse ($T_m = \{T_i\} \cup \text{ch}(T_j)$). Finally, a collapse merges the roots of both trees $T_m = \text{ch}(T_i) \cup \text{ch}(T_j)$.

Each step of the algorithm then consists of picking a pair of trees as well as one of four possible merge operations (there are two absorb possibilities). The pair of trees and merge operation picked are the combination that maximises the likelihood ratio:

$$L(T_m) = \frac{p(\text{leaves}(T_m)|T_m)}{p(\text{leaves}(T_i)|T_i)p(\text{leaves}(T_j)|T_j)} \tag{13}$$

We use the likelihood ratio rather than $p(\text{leaves}(T_m)|T_m)$ because the denominator makes $L(T_m)$ comparable across different choices with trees $T_i$ and $T_j$ of differing sizes (Friedman, 2003; Heller and Ghahramani, 2005).

Choosing a join to construct $T_m$ means that the children of $T_i$ and $T_j$ are related according to $f(\mathcal{D}_m)$ but are sufficiently distinguishable that the two subtrees $T_i$ and $T_j$ should stay separated. Picking absorb ($T_m = \mathrm{ch}(T_i) \cup \{T_j\}$) means that the leaves are similar but there exists some finer distinguishing structure already captured by $T_j$. A collapse is performed when the children of $T_i$ and $T_j$ are indistinguishable and so may be combined and treated similarly. These intuitions in the complexity of the data are reflected in the partition structure of the resulting $T_m$: collapse produces the fewest partitions, join produces the most partitions in $T_m$, while absorbs are in between.

Binary hierarchical clustering algorithms only need to consider the join operation. To be able to construct every possible rose tree the absorb operation is necessary as well. The collapse operation is not technically necessary, however we found that including it allowed us to find better rose trees and we include it as a result. The resulting algorithm has $O(n^2 \log n)$ time and space complexity where $n$ is the number of data points, ignoring complexity due to the particular cluster marginal likelihood $f(\mathcal{D})$ used. The $\log n$ factor is due to searching for the best pair of trees to merge.

**input:** data $\mathcal{D} = \{x^{(1)} \dots x^{(n)}\}$, model $p(x|\theta)$, prior $p(\theta|\beta)$
**initialise:** number of clusters $c = n$, and $T_i = \{x^{(i)}\}$ for $i = 1 \dots n$
**while** $c > 1$ **do**
Find the pair of trees $T_i$ and $T_j$, and merge operation $m$ with the highest likelihood ratio:

$$L(T_m) = \frac{p(\mathrm{leaves}(T_m)|T_m)}{p(\mathrm{leaves}(T_i)|T_i)p(\mathrm{leaves}(T_j)|T_j)}$$

Merge $T_i$ and $T_j$ into $T_m$ using operation $m$
Delete $T_i$ and $T_j$, $c \leftarrow c - 1$
**end while**

Figure 4: Bayesian rose tree algorithm

### 3.1 Hyperparameter optimisation

The hyperparameters $\beta$ of the exponential family distribution for each cluster can be optimised by using gradient ascent. From (3), the gradient of the log likelihood $p(\mathcal{D}|T)$ can easily be computed recursively:

$$\frac{\partial \log p(\mathcal{D}|T)}{\partial \beta} = r_T \frac{\partial \log f(\mathrm{leaves}(T))}{\partial \beta} + (1 - r_T) \sum_{T_i \in \mathrm{ch}(T)} \frac{\partial \log p(\mathrm{leaves}(T_i)|T_i)}{\partial \beta} \tag{14}$$

$$\text{where } r_T = \frac{\pi_T f(\mathrm{leaves}(T))}{p(\mathcal{D}|T)} \tag{15}$$

A similar gradient can be found for $\gamma$, however in our experiments $\gamma$ is optimised using Brent's method.

After optimising these hyperparameters on a particular tree, one option is to find another tree that maximises $p(\mathcal{D}|T)$ using these hyperparameters in an EM-like algorithm: in the E-step, greedily find $T$ then in the M-step find the best hyperparameters, and repeat.

## 4 Results

In this section we present several experiments using BRT. We compare our results to BHC-$\gamma$ and in most cases BHC-DP as well. For BHC-DP we report its marginal likelihood $p(\mathcal{D}|T)$, not the lower bound on the DP mixture, which is $p(\mathcal{D}|T)$ multiplied by a factor that is less than one. Except for the last experiment, the data we used are binary vectors. We assumed $f(\mathcal{D})$ is factorised across dimensions, with dimension $i$ modelled by a Bernoulli distribution with beta$(\alpha_i, \beta_i)$ prior. Integrating out the parameters,

$$f(\mathcal{D}) = \prod_{i=1}^{d} \int f(\mathcal{D}_i|\theta_i) f(\theta_i|\alpha_i, \beta_i) d\theta_i = \prod_{i=1}^{d} \frac{\mathrm{B}(\alpha_i + n_i, \beta_i + N - n_i)}{\mathrm{B}(\alpha_i, \beta_i)} \tag{16}$$

where $d$ is the number of dimensions, $N$ is the number of data points in $\mathcal{D}$, $n_i$ is the number of 1s in dimension $i$ and $B(x, y)$ is the beta function.

**Optimality of tree structure.** The agglomerative scheme described in Section 3 is a greedy algorithm that is not guaranteed to find the optimal tree. Here we compare the trees found by BRT, BHC-$\gamma$ and BHC-DP against the optimal (maximum likelihood) Bayesian rose tree $T^*$ found by exhaustive search. We generated data sets of sizes ranging from 2 to 8, each consisting of binary vectors of dimension 64, from a BRT mixture with randomly chosen rose tree structures. On each of the $N$ data sets we compare the performances in terms of the average log probability of the data assigned by the three greedily found trees $T$ relative to the maximum likelihood Bayesian rose

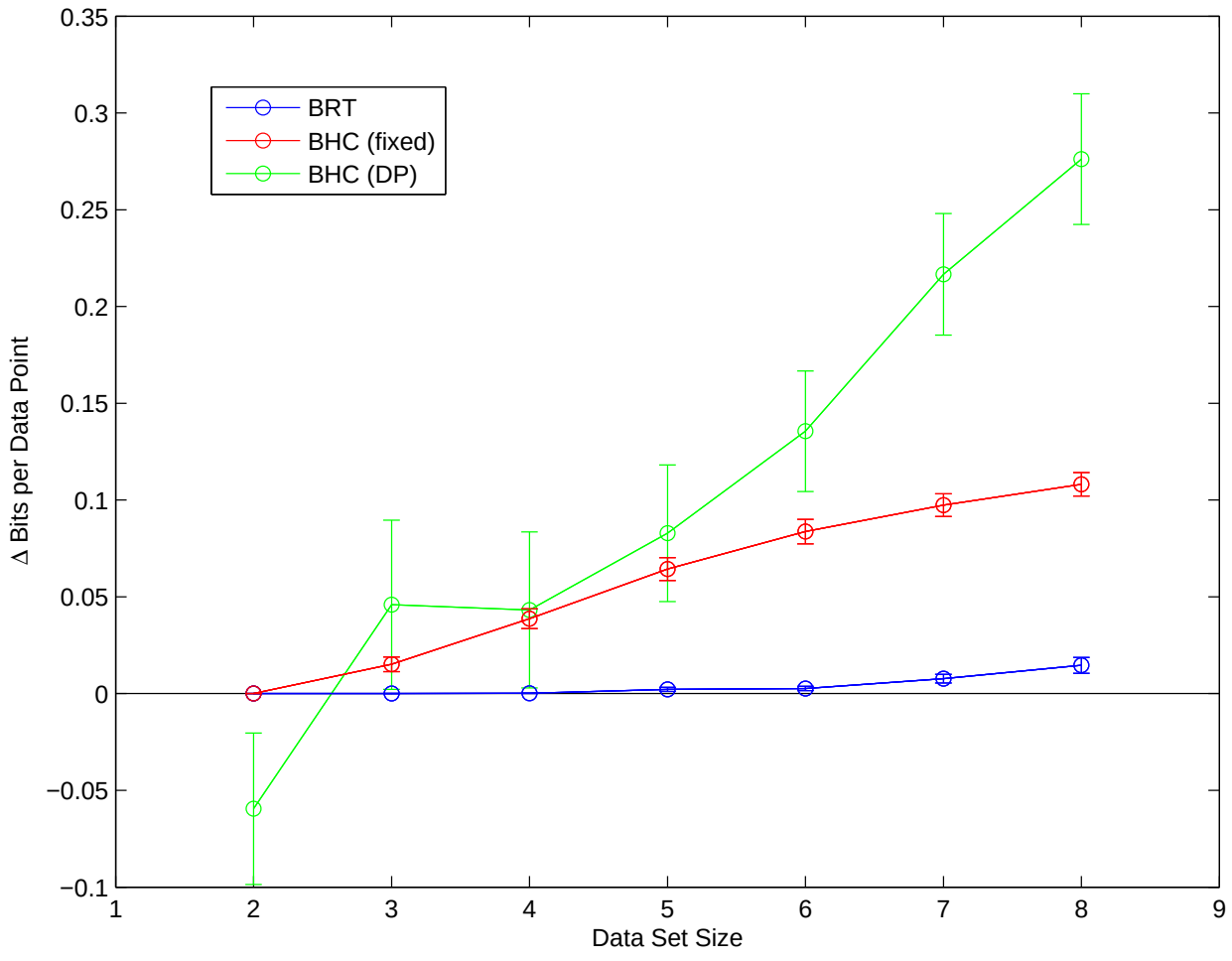

Figure 5: Per data item log probability of trees found greedily by BRT, BHC-$\gamma$ and BHC-DP, relative to the optimal Bayesian rose tree. Error bars are one standard error. Lower in graph is better. Log in base 2.

tree $T^*$,

$$\Delta_l = \frac{1}{lN} \sum_{n=1}^{N} \log_2 p(\mathcal{D}_n|T_n^*) - \log_2 p(\mathcal{D}_n|T_n) \qquad (17)$$

where $l$ is the number of data vectors in the data set. $\Delta_l$ measures the average number of bits required to code for a data vector under $T$, in excess of the same under $T^*$. The results, averaged over 100 data sets per data set size are shown in Figure 5. We see that BRT finds significantly better trees than either BHC algorithms. We also found that BRT frequently finds the optimal tree, e.g. when $l = 8$ BRT found the optimum 70% of the time. Note that when $l = 2$ BHC-DP produced higher log probability than the optimal BRT $T^*$, although it performed significantly worse than BHC-$\gamma$ and BRT for larger $l$. This is because the BHC-DP and BRT models are not nested so BHC-DP need not perform worse than $T^*$.

**Psychological hierarchies.** The data set of Figure 6 is from Cree and McRae (2003) and consists of 60 objects, each with 100 binary attributes (such as used for transportation, has legs, has seeds, is cute). Figure 6 shows the trees found by BRT and BHC-$\gamma$. This figure shows how BRT not only finds simpler, easier to interpret hierarchies than BHC-$\gamma$ but also more probable explanations of the data. The features of the data set include is it ferocious?, does it roar? which only lions and tigers have, whilst they share few attributes common to other animals in this data set: this is why they lie on a distinct branch. For space reasons we did not include BHC-DP, which obtained lower likelihood ($\log p(\mathcal{D}|T) = -1419$) with $468,980,051$ partitions.

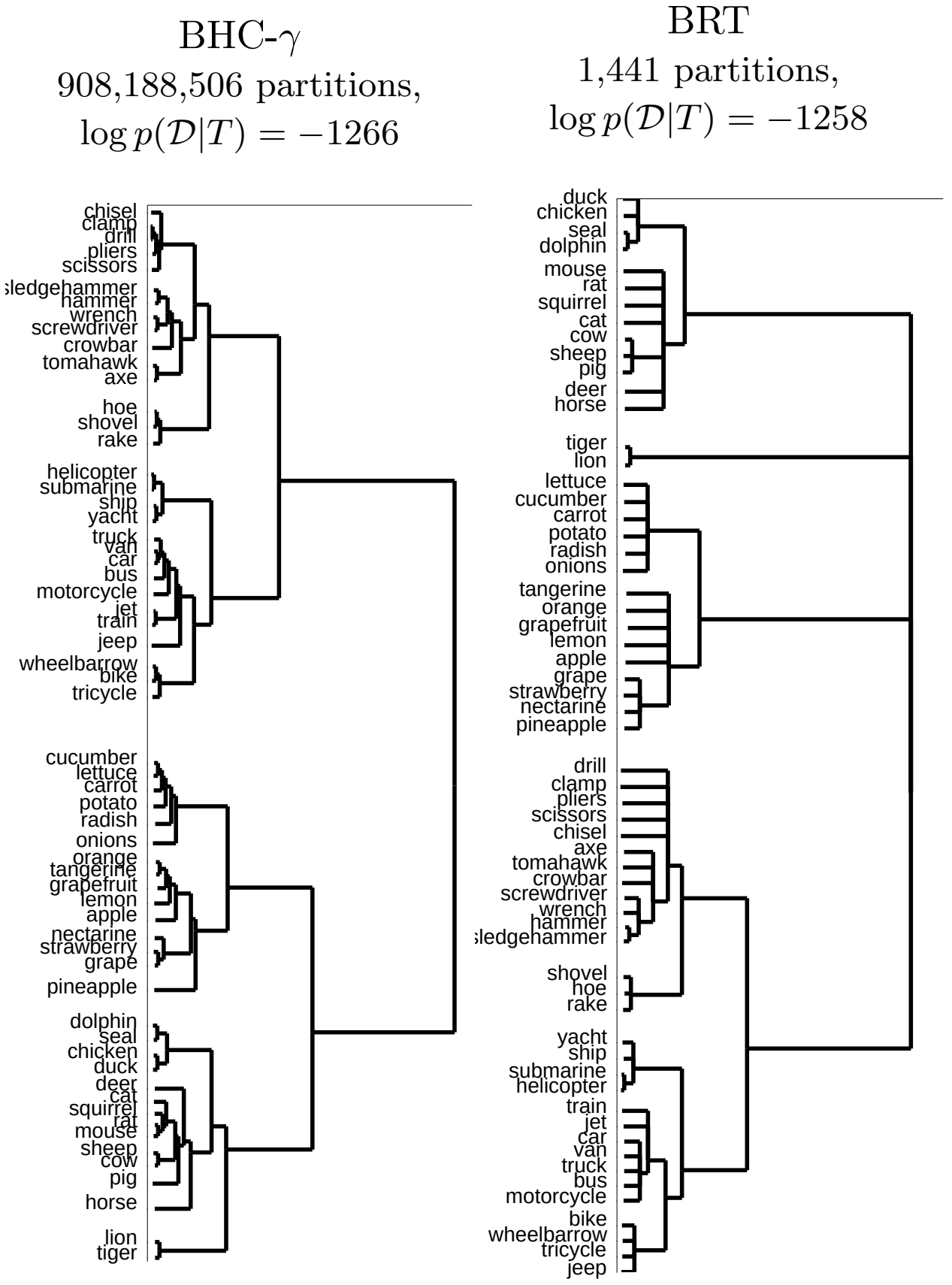

Figure 6: Hierarchies found by BHC-$\gamma$ and BRT on 60 objects (with 100 binary features) data set.

Table 1: Characteristics of data sets.

| Data set | Attributes | Classes | Binarisation |
|---|---|---|---|
| `toy` | 12 | 3 | handcrafted |
| `spambase` | 57 | 2 | zero or non-zero |
| `newgroups` | 485 | 4 | word presence/absence |
| `digits` | 64 | 10 | threshold at 32 |
| `digits024` | 64 | 3 | threshold at 32 |

**Hierarchy likelihoods.** We compared the marginal likelihoods of trees found by BHC-DP, BHC-$\gamma$ and BRT on five other data sets. The characteristics of the data sets are summarised in Table 1. `toy` is a synthetic data set constructed where 1s only appear in three disjoint parts of the binary vector, with each class having 1s in a different part. The hierarchies in Figure 1 are found by BHC-$\gamma$ and BRT on this data set. `spambase` is the UCI repository data set. `newsgroups` is the CMU 20newsgroups data set restricted to the news groups `rec.autos`, `rec.sport.baseball`, `rec.sport.hockey`, and `sci.space`, constructed using Rainbow (McCallum, 1996). `digits` is a subset of the CEDAR Buffalo digits data set, and `digits024` is the same data set with only samples corresponding to the digits 0, 2, and 4. Each data set consists of

Table 2: Log likelihoods and standard errors

| Data set | BHC-DP | BHC-$\gamma$ | BRT |
|---|---|---|---|
| toy | $-192 \pm 0$ | $-169 \pm 0$ | $\mathbf{-166} \pm 0$ |
| spambase | $-2354 \pm 4.7$ | $-2000 \pm 4.5$ | $\mathbf{-1991} \pm 4.5$ |
| digits024 | $-4154 \pm 5.2$ | $-3759 \pm 4.6$ | $\mathbf{-3748} \pm 4.6$ |
| digits | $-4429 \pm 3.3$ | $-3966 \pm 3.1$ | $\mathbf{-3954} \pm 3.1$ |
| newsgroups | $-11602 \pm 104$ | $-10833 \pm 106$ | $-10827 \pm 105$ |

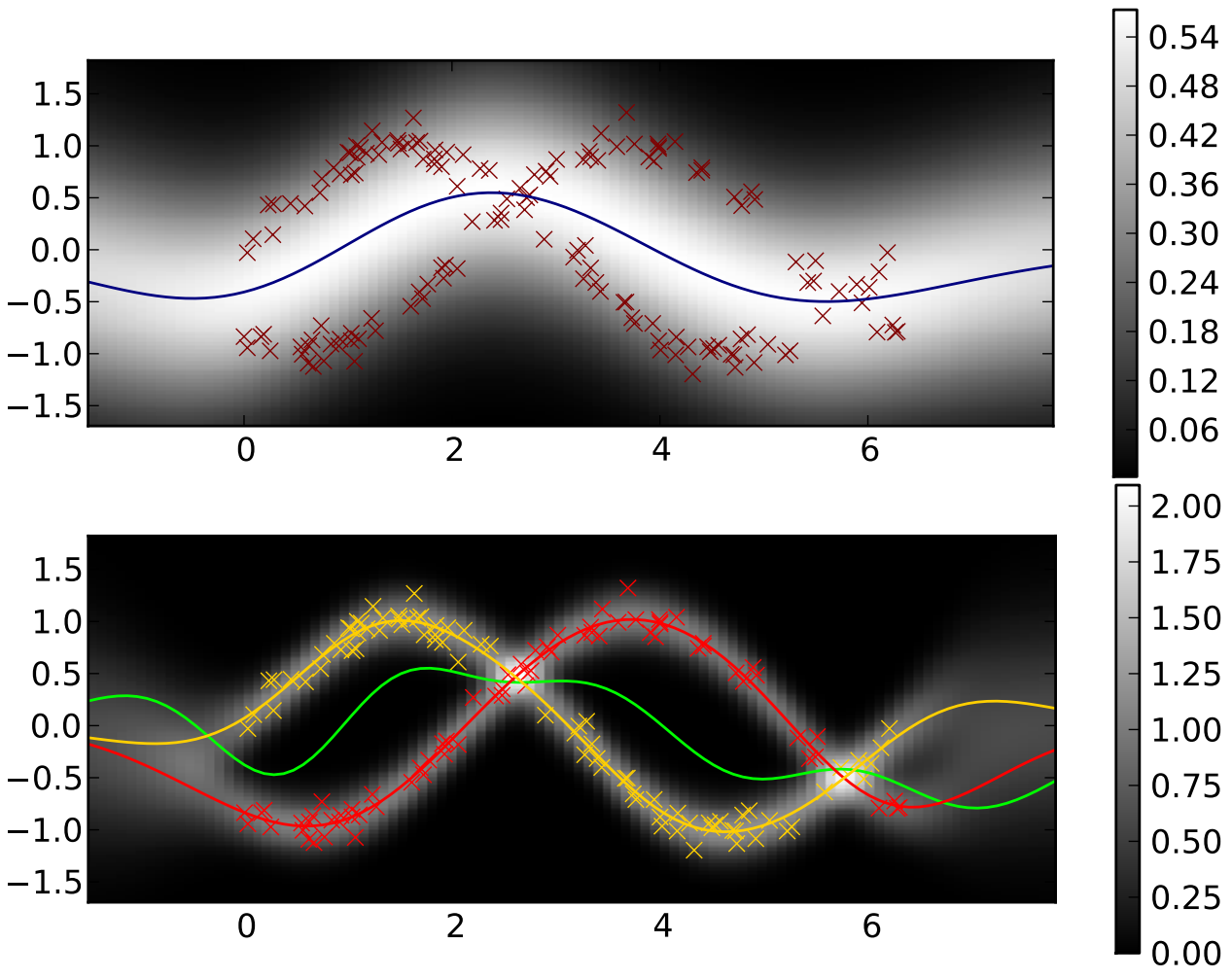

Figure 7: A Gaussian process expert (top) with marginal log likelihood: −1037, on synthetic data of two interlaced curves. Observations are crosses, and line is the posterior mean function pf the GP. Background grey scale indicates the predictive density of the GP (scale on left of top plot). On the bottom, a BRT mixture of GP experts (bottom) with marginal log likelihood: 59. Background grey scale indicates corresponding density $p(y|x, \mathcal{D})$ (as defined in Heller (2008)) via scale on corresponding left of plot. Each internal node is uniquely coloured, and its posterior mean function drawn in that colour. Data items (crosses) are coloured according to their parent node.

120 data vectors split equally among the classes, except for toy which has only 48 data vectors. When the original data sets are larger the 120 data vectors are subsampled from the original. The log likelihoods $\log p(\mathcal{D}|T)$ of the found trees are shown in Table 2. In all of our experiments we found that BRT finds higher log likelihoods than BHC-$\gamma$, which in turn finds higher log likelihoods than BHC-DP.

**BRT mixture of Gaussian process experts.** Rasmussen and Ghahramani (2002) proposed a DP mixture of Gaussian process (GP) experts where a data set is partitioned, via the DP mixture, into clusters each of which is modelled by a GP. Such a model can be used for nonparametric density regression, where a full conditional density over an output space is estimated for each value of input. This allows generalisation of GPs allowing for multi-modality and non-stationarity. The original model in Rasmussen and Ghahramani (2002) had mixing proportions which do not depend on input values; this was altered in the paper in an ad hoc manner using radial basis function kernels. Later Meeds and Osindero (2006) extended the model by using a full joint distribution over both inputs and outputs, allowing for properly defined input dependent mixing proportions. With both approaches MCMC sampling was required for inference, which might be slow in convergence.

Here we consider using Bayesian rose trees instead. Let the data points $\mathcal{D} = \{(x_i, y_i)\}_{i=1}^N$ where $x_i$ is the input and $y_i$ is the output. The joint distribution of each cluster is modelled using a Gaussian over the inputs and a GP over the outputs given the inputs:

$$f(\mathcal{D}) = f(\{x_i \in \mathcal{D}\}) f(\{y_i \in \mathcal{D}\}|\{x_i \in \mathcal{D}\}) \tag{18}$$

where $f(\{x_i \in \mathcal{D}\})$ is the marginal probability of the inputs under a Gaussian with a conjugate Gaussian-inverse-Wishart prior, and $f(\{y_i \in \mathcal{D}\}|\{x_i \in \mathcal{D}\})$ is the marginal probability of the outputs given inputs under a GP. We used a squared exponential kernel for the GPs, with length scale, signal variance, and noise variance optimised by gradient ascent in log likelihood.

Once the rose tree has been constructed, given a new input $x$ the posterior probabilities over clusters can be computed as in Heller (2008). The predictive distribution over output $y$ is then a mixture of Gaussians, with mixing proportions given by the posterior over clusters, while each Gaussian is the distribution of $y$ under the GP in the corresponding cluster, conditional on the other input/output pairs in the cluster.

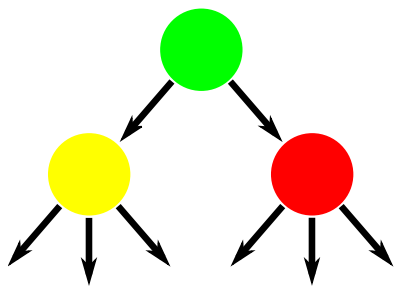

Figure 8: Bayesian rose tree of Gaussian process experts shown of figure 7.

Figure 7 shows the BRT estimated conditional densities as well as the conditional densities under a single GP, on a synthetic multi-modal data set. BRT constructs the rose tree in Figure 8. Each internal nodes is represented in Figure 7 with the mean of its GP component. Two of internal nodes correspond to the two modalities in the data set (these are coloured yellow and red). The third node (the root; green) is the parent of these nodes. The BRT mixture of GP experts has a higher likelihood than that of the GP for this data set.

## 5 Discussion

We have described a model and developed an algorithm for performing efficient, non-binary hierarchical clustering. Our Bayesian rose tree approach is based on model selection: each tree is associated with a mixture of partitions of the data set, and a greedy agglomerative algorithm finds trees that have high marginal likelihood under the data.

Bayesian rose trees are a departure from the common binary trees for hierarchical clustering. The flexibility implied by a mixture model over partitions with tree consistency is used in Bayesian rose trees to allow mixture models with fewer components, and thus simpler explanations of the data, than those afforded by BHC. The BRT mixture proportions are designed so that simpler models which explain the data are favoured over more complicated ones: this is in contrast to BHC-DP where forced binary merges create extra, spurious structure which is not supported by the data.

We have demonstrated in our experiments that this algorithm finds simple models which explain both synthetic and real-world data. On all data sets considered, our Bayesian rose tree algorithm found a rose tree with higher marginal likelihood under the data than Bayesian hierarchical clustering (BHC) (Heller and Ghahramani, 2005). We built BRTs using two likelihood models, a beta-Bernoulli and a Gaussian process expert. In both cases the model yielded reasonable mixtures of partitions of the data.

Our use of BRT for nonparametric conditional density estimation is a proof of concept. BRT offers an attractive means of fitting a mixture of GP experts compared to sampling (Rasmussen and Ghahramani, 2002; Meeds and Osindero, 2006): with sampling one is never sure when the stationary distribution is attained, while the BRT algorithm is guaranteed to terminate after a greedy pass through the data set, constructing a reasonably good estimate of the conditional density. Note however that the run time of the current algorithm is $O(n^5 \log n)$ where $n$ is the number of data points. The additional $O(n^3)$ factor is due to the unoptimised GP computations. An interesting future project would be to make the computations more efficient using recent advanced approximations.

There are many related methods to the Bayesian rose trees presented here. These include recent methods based on probabilistic or Bayesian hierarchical clustering which operate agglomeratively (Friedman, 2003; Heller and Ghahramani, 2005; Teh et al., 2008), but are restricted to binary merges. Others (Neal, 2003; Roy et al., 2007; Teh et al., 2008) describe Bayesian nonparametric priors and propose to compute posteriors over binary trees. These are desirable since they result in uncertainty over tree structures, but are less computationally efficient as a result. Nonparametric priors also exist over rose trees (Pitman, 1999) and could be used for hierarchical clustering as well. This is an avenue for future exploration.